\documentclass[10pt,twocolumn,letterpaper]{article}
\usepackage[pagenumbers]{cvpr}
\usepackage[T1]{fontenc}
\usepackage{microtype}
\usepackage{tabularx,colortbl}
\definecolor{palimpsestMocha}{HTML}{A47864}
\definecolor{palimpsestSubhead}{HTML}{775344}
\newcommand{\palimpsestparagraph}[1]{%
  \paragraph{\texorpdfstring{\textcolor{palimpsestSubhead}{#1}}{#1}}}
\definecolor{cvprblue}{rgb}{0.21,0.49,0.74}
\usepackage[draft,pagebackref]{hyperref}
\def\confName{CVPR}
\def\confYear{2027}
\hypersetup{pdfauthor={Tianjun Gu, Shao Zeng, Ruixin Zhang, Chenghua Gong, Haoran Li, Haonan Wang, Yuan Xie, Jingyu Gong, Jingyun Zhang, Jun Wang, Rizen Guo, Lizhuang Ma, Shouhong Ding, Xin Tan}}

\title{\texorpdfstring{\protect\raisebox{-0.18em}{\protect\includegraphics[height=1.15em,trim=0 36 0 36,clip]{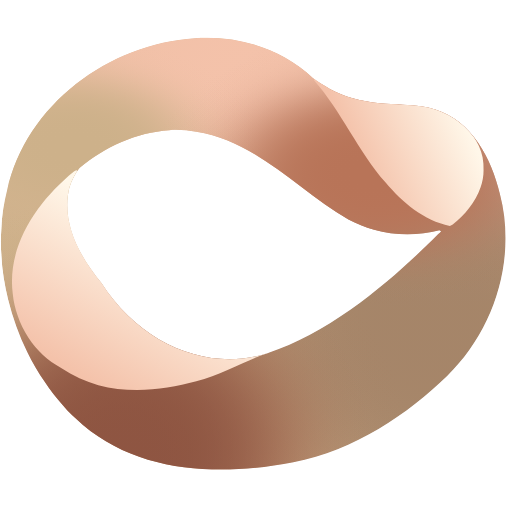}}\enspace}{}\texorpdfstring{\textcolor{palimpsestMocha}{Palimpsest}}{Palimpsest}: Spatial Belief Improvement through Self-Evolving Agents}
\author{\normalsize Tianjun Gu\textsuperscript{1}, Shao Zeng\textsuperscript{2}, Ruixin Zhang\textsuperscript{2}, Chenghua Gong\textsuperscript{6}, Haoran Li\textsuperscript{1}, Haonan Wang\textsuperscript{2}, Yuan Xie\textsuperscript{1,5}\\
\normalsize Jingyu Gong\textsuperscript{1}, Jingyun Zhang\textsuperscript{4}, Jun Wang\textsuperscript{4}, Rizen Guo\textsuperscript{4}, Lizhuang Ma\textsuperscript{1}, Shouhong Ding\textsuperscript{2}, Xin Tan\textsuperscript{1,3}\thanks{Corresponding author: \texttt{xtan@cs.ecnu.edu.cn}.}\\[2pt]
\small \textsuperscript{1}East China Normal University\quad \textsuperscript{2}Tencent Youtu Lab\quad \textsuperscript{3}Shanghai AI Laboratory\\
\small \textsuperscript{4}WeChat Pay Lab 33, Tencent\quad \textsuperscript{5}Shanghai Innovation Institute\quad \textsuperscript{6}University of Science and Technology of China}

\makeatletter
\AddToHook{cmd/@maketitle/after}{%
  \vspace{-32pt}%
  \noindent
\begin{minipage}{\textwidth}
\centering
\includegraphics[width=0.775\textwidth]{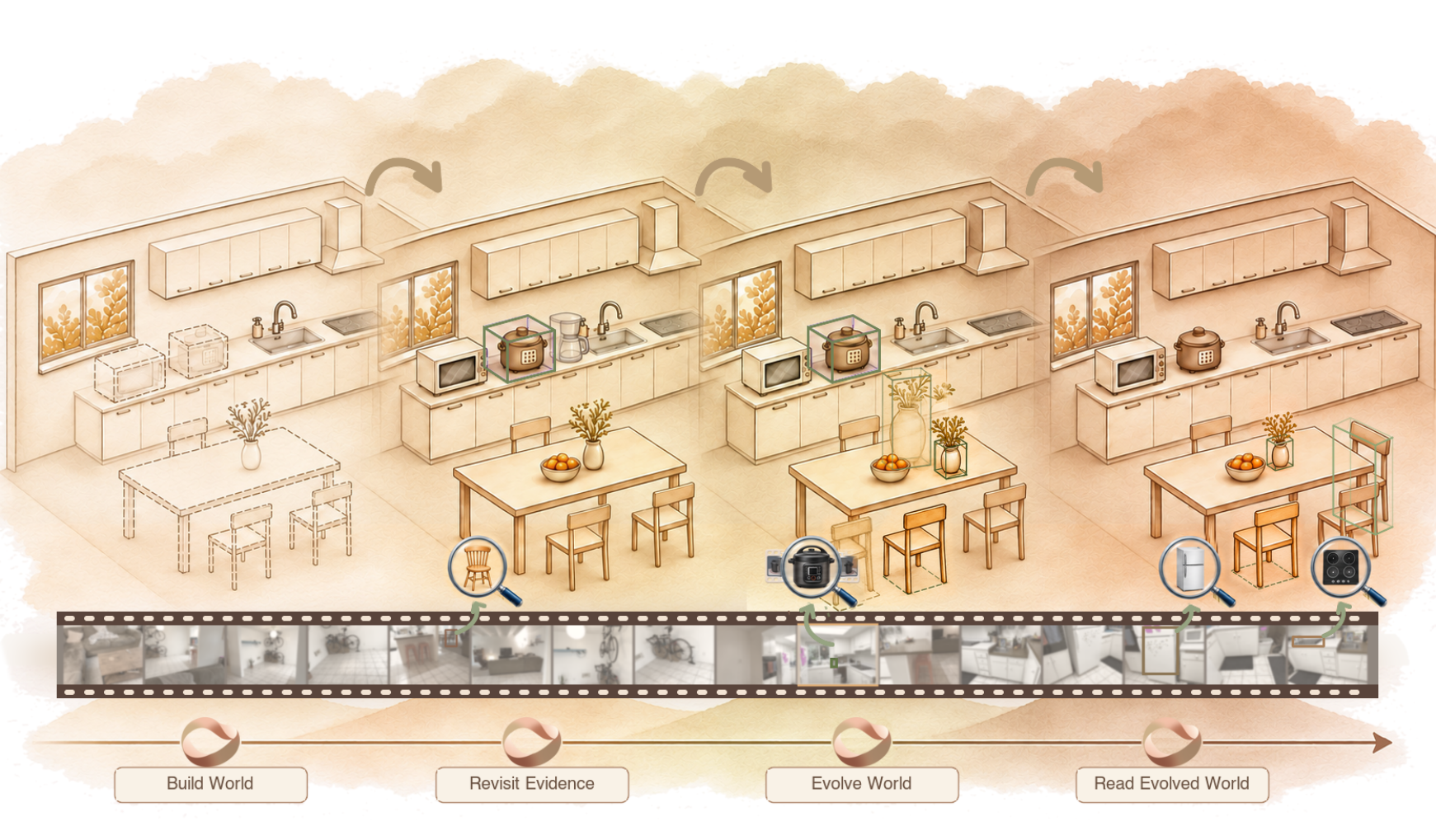}
\captionof{figure}{\textbf{Revising a spatial world from video evidence.}
Palimpsest builds an initial world from video, revisits relevant observations to check uncertain records, and writes accepted revisions back for spatial programs to use.}
\label{fig:teaser}
\end{minipage}\par%
\vspace{8pt}%
}
\makeatother

\DeclareCaptionType{algorithm}[Algorithm][List of Algorithms]

\usepackage{placeins}
\usepackage{stfloats}
\definecolor{suppHeader}{HTML}{EADCD1}
\definecolor{suppHighlight}{HTML}{F1E9E3}
\newcommand{\suppparagraph}[1]{\paragraph{\textcolor{palimpsestSubhead}{#1}}}
\newenvironment{suppprompt}[1]{\par\medskip\noindent
  {\small\bfseries\color{palimpsestSubhead}#1}\par\nopagebreak\small\noindent}
  {\par\medskip}

\begin{document}
\maketitle

\begin{abstract}
Spatial reasoning from video requires integrating partial observations across time and viewpoints. Missing objects and incorrect labels in the initial world can affect multiple answers about the same scene. We introduce Palimpsest, an agent that builds an executable spatial world before receiving any questions. The world links entity names and geometry to their supporting video observations. When an entity has weak support or a question exposes an unresolved reference, the agent revisits relevant views and writes accepted revisions back to the world. Spatial programs then compute answers from the revised records. Across 380 ReVSI scenes, Palimpsest scores 67.2, a gain of 19.9 points over direct video answering with the same model backbone. In a separate test of missing-object recovery, revisions triggered only by other questions improve the score by 7.21 points over the world before recovery. This amounts to 87\% of the gain obtained when all questions are allowed to trigger recovery.
\end{abstract}

\vspace*{\fill}
\noindent
\begin{minipage}{\columnwidth}
\centering
\includegraphics[width=\columnwidth,height=190pt,keepaspectratio]{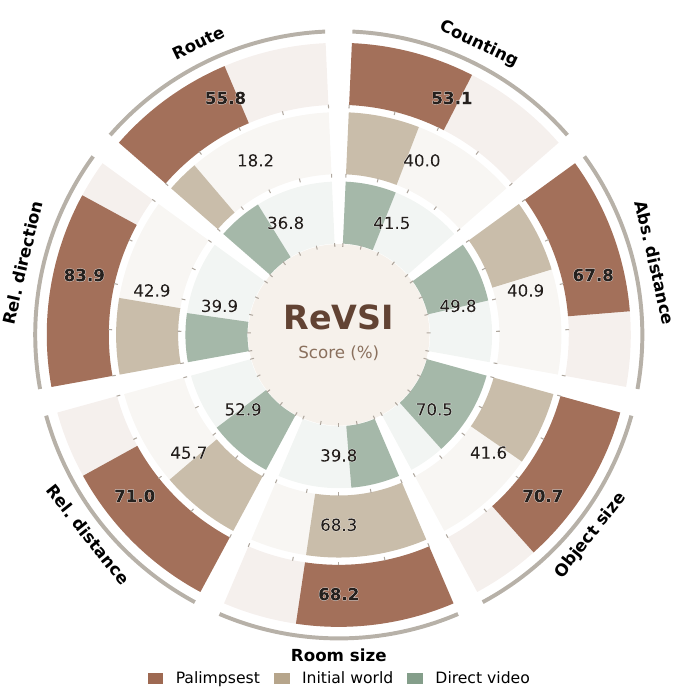}
\captionof{figure}{\textbf{Spatial reasoning across seven ReVSI tasks.}
The rings show scores for Palimpsest, the initial world, and direct video answering, from outside to inside.}
\label{fig:revsi-overview}
\end{minipage}
\par
\newpage

% Prefer a substantial final line in justified paragraphs.
\setlength{\parfillskip}{0pt plus 0.58\columnwidth}
\section{Introduction}

Spatial reasoning from egocentric video requires integrating observations across time and viewpoints. As the camera moves, different parts of a scene come into view, and the same object may reappear from a new angle or under different occlusions. Benchmarks such as VSI-Bench test whether models can recover spatial information from these partial observations~\cite{yang2025thinking}. Explicit object maps organize observations in shared scene coordinates~\cite{gu2024conceptgraphs}. Tool-using agents such as S-Agent can also gather evidence on demand and integrate it into scene memory~\cite{dai2026sagent}. Together, these approaches let an agent assemble and consult a spatial representation.

A spatial world can contain the right object and still give the wrong answer. A program may need the distance to a multi-cooker, for example, while the reconstructed world lists that object under a different name. The unresolved reference gives the agent a reason to revisit the video and check what it recorded. We call this feedback \emph{self-evolution}: gaps exposed by inspecting or querying the world guide evidence checks, and the resulting revisions persist in the world for later computations. \Cref{fig:teaser} illustrates this process.

We introduce \textbf{Palimpsest}, a self-evolving agent for spatial reasoning from video. Before receiving any questions, Palimpsest builds an executable spatial world that links entity records to their names, geometry, and supporting observations. Weak support in an entity record or an unresolved reference in a question can trigger a return to relevant views. The agent checks the affected records against those observations and writes accepted revisions back to the world. Spatial programs use those records to compute distances, directions, and counts.

On 380 ReVSI scenes~\cite{revsi2026}, the full Palimpsest system scores \textbf{67.2}, compared with 47.32 for direct question answering from 32 video frames. A separate experiment tests whether objects recovered for one set of questions help answer another.

Our contributions are as follows.
\begin{itemize}
    \item We formulate self-evolving spatial reasoning as revising a shared world when gaps in its records prompt further inspection of the video.
    \item We develop Palimpsest, which builds an executable spatial world from video before seeing any questions. It answers questions by running spatial programs on that world.
    \item Across all 380 ReVSI scenes, Palimpsest achieves an overall score of \textbf{67.2}, a gain of \textbf{19.9 points} over direct question answering from 32 video frames, which scores 47.32. Across those scenes, missing-object recovery triggered by other questions gains 7.21 points over the world before recovery.
\end{itemize}

\section{Related Work}

\paragraph{Spatial reasoning from video.}
VSI-Bench evaluates spatial reasoning from video, and ReVSI revises its annotations and aligns questions with the visual input available to the model~\cite{yang2025thinking,revsi2026}. VSI-Super-Wild extends evaluation to long videos recorded in diverse real-world settings and identifies insufficient state updates and instance confusion among observed failures~\cite{gu2026vsiwild}. EscherVerse examines physical dynamics and action goals~\cite{escherverse2026}. For spatial inference, Thinking with Spatial Code brings object and geometry cues into the model's representation~\cite{chen2026spatialcode}. S-Agent uses spatial tools and scene memory to gather and integrate evidence during reasoning~\cite{dai2026sagent}, while SpatialClaw allows an agent to compose perception and geometry operations in a stateful Python kernel~\cite{cho2026spatialclaw}. Palimpsest studies how evidence gathered during reasoning can revise the scene records used to answer spatial questions.

\paragraph{Spatial representations and scene memory.}
ConceptFusion integrates multimodal features into a queryable 3D map, and ConceptGraphs organizes objects and relations into an open-vocabulary scene graph~\cite{jatavallabhula2023conceptfusion,gu2024conceptgraphs}. Hydra constructs and optimizes hierarchical scene graphs online~\cite{hughes2022hydra}. In embodied navigation, DORAEMON combines semantic observations with a topological map to support decisions over time~\cite{doraemon2025}. For video understanding, the memory-augmented VideoAgent stores event descriptions alongside object tracks~\cite{fan2024memoryvideoagent}. Palimpsest links its entity records to the video observations that support them, so the same records can be checked against evidence and used in spatial computations.

\paragraph{State revision during inference.}
Self-Refine uses model-generated feedback to revise an initial output~\cite{madaan2023selfrefine}, and Reflexion stores verbal feedback to guide subsequent trials~\cite{shinn2023reflexion}. In model-based control, WorldCoder uses interaction experience to revise an executable model of environment transitions~\cite{tang2024worldcoder}. For visual reasoning, VideoAgent retrieves relevant moments from long videos~\cite{wang2024videoagent}, and VisProg and ViperGPT compose visual operations through programs~\cite{gupta2023visprog,suris2023vipergpt}. Palimpsest uses retrieved visual evidence to revise the scene records its spatial programs read. Weak visual support for an entity or an unresolved reference in a question can trigger a check. Accepted changes are written back to the world for subsequent computations.

\section{Method}

\begin{figure*}[t]
\centering
\includegraphics[width=\textwidth]{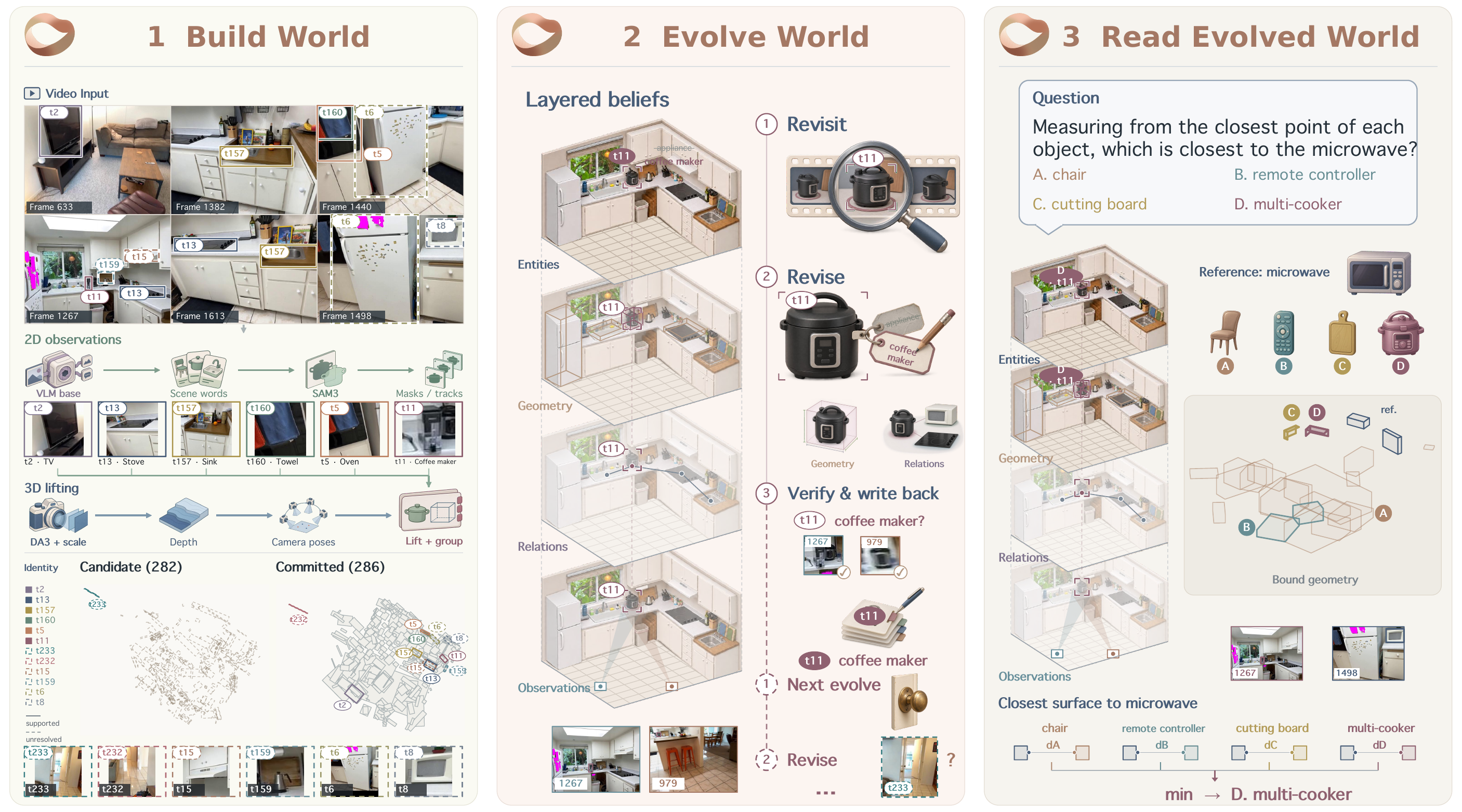}
\caption{\textbf{Building, evolving, and reading the spatial world.}
\textbf{Build.} The agent groups video observations into entities and records their geometry before seeing any questions.
\textbf{Evolve.} The agent revisits marked views and adds a binding from \emph{multi-cooker} to the existing entity \texttt{t11}.
\textbf{Read.} The reader uses the existing box for \texttt{t11} to compare the options' distances to the microwave. The comparison selects option D as the closest to the microwave.}
\label{fig:method}
\end{figure*}

Palimpsest builds a spatial world from video, revisits the video when a record needs checking, and computes answers from the revised world. \Cref{fig:method} shows how these stages share the same representation. The links from entities to their supporting observations let the agent revisit the evidence for a specific judgment, and revisions to those entities change the records available to the reader.

\subsection{The executable spatial world}
\label{sec:spatial-world}

Palimpsest maintains one world per scene. The world stores video frame identifiers and camera parameters, mask observations indexed by frame, and entities assembled from those observations. An entity identifier links its supporting observations to its estimated geometry, including its 3D center and extent, and to attributes such as its candidate names. Thus, a reader can use an entity's geometry for a distance calculation, and the agent can return to the frames that support that entity when its name or existence needs checking.

Each entity has a status that governs whether it is eligible for ordinary readout. A committed entity is eligible, while an entity retained as a hypothesis keeps its observations available for later checks or promotion. The world stores revision entries that record changes such as promotion, withdrawal, or renaming, along with the evidence or prior values recorded by the corresponding operation. Spatial programs derive distances and directions from the current entity records. Updating a name binding changes which geometry a program can access through that name, and withdrawing an entity changes the set of records eligible for computation.

\subsection{Build the world from video}

The initial world is built before any question is seen. It gives the agent a scene-wide set of entity hypotheses and preserves the observations needed to check them later.

\palimpsestparagraph{Observations from video.}
We combine the object names proposed by the vision-language model from 32 video frames with a fixed vocabulary drawn from ScanNet200~\cite{rozenberszki2022scannet200}. A vocabulary filter removes terms such as object parts, materials, and generic regions. SAM~3~\cite{carion2025sam3} checks for the presence of each retained term and tracks detected objects across the sampled video frames. In parallel, Depth Anything~3~\cite{lin2025da3} estimates depth and camera parameters. We calibrate scale with DA3METRIC-LARGE and back-project each retained mask pixel $\mathbf u=(u,v,1)^\top$ into the scene:
\begin{equation}
\mathbf x_t(\mathbf u)=R_t\bigl(d_t(\mathbf u)K_t^{-1}\mathbf u\bigr)+\mathbf t_t .
\end{equation}
Here $t$ indexes the frame, $(u,v)$ are image coordinates in pixels, and $d_t(\mathbf u)$ is the depth at that pixel in meters. $K_t$ is the camera intrinsic matrix. The camera-to-world rotation $R_t$ and translation $\mathbf t_t$ map the back-projected point to the world position $\mathbf x_t(\mathbf u)$ in meters. Each lifted observation keeps its frame and mask reference, so an entity assembled in 3D can still be checked against its original image evidence.

\palimpsestparagraph{From observations to entities.}
We initialize one candidate per track. Tracks with different names can be merged when their masks repeatedly overlap; tracks with the same name can be merged when their 3D centers are close. We reject a merge if the resulting group would contain spatially disjoint masks with the same name in a shared frame. For each resulting entity, we estimate a median-based 3D center and a robust extent from its lifted observations, and use its identifier to store names and observation references.

\palimpsestparagraph{Initial commitment.}
Let $O_e$ be the frames in which entity $e$ has a mask observation and $V_e$ the frames in which its center is expected to be visible. The builder computes coverage $c_e=|O_e|/|O_e\cup V_e|$ and commits an entity if $|O_e|\geq 2$ or $c_e\geq 0.5$. Smaller fragments split from a track group require at least three observed frames. Entities that fail this rule remain in the world as hypotheses. Commitment is therefore an initial decision about which records the reader can use. By revisiting the video, the agent can check both the existence and the name of a committed entity.

\subsection{Evolve the world}

The initial commitment rule uses the support available during construction. Once the world is built, Palimpsest can revisit the video to check an entity's existence or name. A question can expose a different problem when its referring phrase has no usable binding. In either case, the agent starts from the current world, retrieves evidence for the record in question, and writes accepted changes back to the scene state.

\palimpsestparagraph{Check existing entities.}
The agent prioritizes entities with few observations, conflicting names, or missing detections in views where the entity is expected to be visible. Both existence and name checks use crops from the entity's linked mask observations. Projected-view fallback is reserved for later phrase-specific checks with insufficient observed views. It favors views that add a different viewing direction and tracks views already judged. The vision-language model evaluates the selected evidence. For an existence check, let $n_+$ and $n_-$ count affirmative and negative judgments on the selected views. The check confirms existence when $n_+-n_-\geq 2$ and rejects it when $n_--n_+\geq 2$. Name checks can remove rejected names or retain a newly supported one. A withdrawn entity returns to hypothesis status, with its original observations available for later checks. The ledger records the revisions and the judgments used in each round.

\palimpsestparagraph{Recover missing referents.}
When a phrase fails to bind, Palimpsest can use it to prompt segmentation and tracking, then check the resulting world against the video. It also checks synonym-based bindings on entity crops. If a phrase still lacks a committed referent, the agent can promote a hypothesis with an exact name or use a new detection to seed an entity. Recovery assumes the question names a real object. A plausible detection can therefore still produce a wrong binding.

A missing phrase may also refer to an entity already in the world under a different name. The model shortlists plausible names and uses marked views to select among multiple candidates; a single available candidate is selected directly. In \cref{fig:method}, the agent adds \emph{multi-cooker} as a binding to entity \texttt{t11}, which the world already calls \emph{coffee maker}. The ledger records the association, and the entity retains its existing geometry. A spatial program can then use that geometry when asked about the multi-cooker.

\setcounter{dbltopnumber}{1}
\begin{table*}[!t]
\centering
\begingroup%
\definecolor{revsiprimary}{HTML}{A47864}%
\definecolor{revsiink}{HTML}{775344}%
\definecolor{revsiline}{HTML}{D6CDC7}%
\definecolor{revsilight}{HTML}{F1E9E3}%
\definecolor{revsiours}{HTML}{E5D3C7}%
\definecolor{revsiheader}{HTML}{EADCD1}%
\definecolor{revsibestfill}{HTML}{D5B09A}%
\definecolor{revsisecondfill}{HTML}{F1E1D6}%
\newcommand{\revsibest}[1]{\cellcolor{revsibestfill}\textbf{#1}}%
\newcommand{\revsisecond}[1]{\cellcolor{revsisecondfill}#1}%
\arrayrulecolor{revsiline}%
\setlength{\tabcolsep}{2.4pt}%
\setlength{\aboverulesep}{0pt}%
\setlength{\belowrulesep}{0pt}%
\renewcommand{\arraystretch}{1.14}%
\fontsize{7.2}{8.4}\selectfont%
\begin{tabularx}{\linewidth}{lXlcccccccc}%
\arrayrulecolor{revsiprimary}\toprule[1.1pt]\arrayrulecolor{revsiline}%
\rowcolor{revsiheader} & & & & \multicolumn{4}{c}{\color{revsiink}Numerical (MRA)} & \multicolumn{3}{c}{\color{revsiink}Multiple choice (Acc.)} \\%
\rowcolor{revsiheader} \textcolor{revsiink}{\textbf{Method}} & \textcolor{revsiink}{\textbf{Backbone / model}} & \textcolor{revsiink}{\textbf{VLM input}} & \textcolor{revsiink}{\textbf{Avg.}} & \textcolor{revsiink}{\textbf{Count}} & \textcolor{revsiink}{\textbf{Abs. dist.}} & \textcolor{revsiink}{\textbf{Obj. size}} & \textcolor{revsiink}{\textbf{Room size}} & \textcolor{revsiink}{\textbf{Rel. dist.}} & \textcolor{revsiink}{\textbf{Rel. dir.}} & \textcolor{revsiink}{\textbf{Route}} \\%
\midrule%
\rowcolor{revsilight}\multicolumn{11}{l}{\color{revsiink}\textbf{General VLMs}} \\%
GPT-5.2\,\textsuperscript{R} & GPT-5.2 & 64 & 50.9 & 56.2 & 41.5 & 73.9 & 63.0 & 48.4 & 34.9 & 38.2 \\%
Gemini 3 Flash\,\textsuperscript{R} & Gemini 3 Flash & 1 FPS & 57.6 & \revsibest{65.7} & 53.1 & \revsisecond{77.6} & 52.8 & 64.6 & 47.9 & 41.8 \\%
Gemini 3 Pro\,\textsuperscript{R} & Gemini 3 Pro & 1 FPS & 60.9 & 60.1 & 54.7 & \revsibest{79.3} & 51.9 & 68.1 & 56.0 & \revsibest{56.4} \\%
Qwen3-VL\,\textsuperscript{R} & 8B-Instruct & 64 & 49.1 & 40.4 & 52.3 & 69.0 & 45.1 & 57.1 & 39.5 & 40.5 \\%
Qwen3-VL\,\textsuperscript{R} & 32B-Instruct & 64 & 53.3 & 46.9 & 65.0 & 70.4 & 55.8 & 53.8 & 34.0 & 47.3 \\%
InternVL3.5\,\textsuperscript{R} & 38B & 64 & 54.1 & 43.8 & 60.6 & 70.2 & 58.4 & 57.4 & 45.9 & 42.7 \\%
\midrule%
\rowcolor{revsilight}\multicolumn{11}{l}{\color{revsiink}\textbf{Spatial models}} \\%
VLM-3R\,\textsuperscript{S} & LLaVA-Video-7B & 32 & 50.1 & 41.6 & 61.6 & 64.8 & 52.5 & 46.5 & 49.5 & 34.1 \\%
GeoThinker\,\textsuperscript{S} & Qwen3-VL-8B & 32 & 54.0 & 37.1 & \revsisecond{71.5} & 66.0 & 53.0 & 61.9 & 47.5 & 41.5 \\%
SpatialStack\,\textsuperscript{S} & Qwen3.5-4B & 32 & 54.1 & 36.3 & 64.6 & 63.2 & 54.4 & 62.9 & 47.7 & 49.6 \\%
SpatialSpeak\,\textsuperscript{S} & Qwen3-VL-4B & 32 & \revsisecond{62.8} & \revsisecond{64.9} & 70.2 & 62.5 & 66.0 & \revsibest{71.9} & 51.1 & 52.7 \\%
Video-ORA\,\textsuperscript{V} & Qwen3.5-9B & 64 & 55.8 & 46.4 & 65.3 & 70.4 & 50.2 & 62.3 & 47.4 & 48.8 \\%
Video-ORA\,\textsuperscript{V} & Qwen3.5-9B & 128 & 58.2 & 48.8 & 69.1 & 72.7 & 51.1 & 66.3 & 49.9 & 49.8 \\%
Render2Reason\,\textsuperscript{T} & InternVL3.5 (+ geo.) & 64 & 51.4 & 38.4 & 60.0 & 60.5 & 52.3 & 42.9 & 50.3 & 55.2 \\%
Render2Reason\,\textsuperscript{T} & Qwen3-VL (+ geo.) & 64 & 54.5 & 42.2 & \revsibest{74.0} & 67.3 & 58.5 & 41.3 & 51.6 & 46.6 \\%
Cambrian-S\,\textsuperscript{V} & Qwen2.5-7B & 128 & 49.1 & 48.4 & 60.5 & 65.5 & 46.7 & 37.1 & 48.5 & 37.0 \\%
Cambrian-P\,\textsuperscript{F} & Qwen2.5-7B & 128 & 52.0 & 41.4 & 68.3 & 66.7 & 45.9 & 40.2 & 48.7 & 53.1 \\%
GeoThinker-F\,\textsuperscript{F} & Qwen3-VL-8B & 128 & 50.6 & 36.4 & 65.2 & 66.1 & 56.2 & 45.9 & 46.1 & 38.5 \\%
Cambrian-BEV\,\textsuperscript{B} & 0.5B & 64 & 45.0 & 37.6 & 41.7 & 46.3 & 60.2 & 34.6 & 44.3 & 50.2 \\%
Cambrian-BEV\,\textsuperscript{B} & 0.5B & 128 & 44.7 & 38.1 & 41.1 & 46.8 & 58.7 & 34.7 & 43.8 & 49.5 \\%
\midrule%
\rowcolor{revsilight}\multicolumn{11}{l}{\color{revsiink}\textbf{Tool-assisted methods}} \\%
CROSS\,\textsuperscript{C} & Qwen3-VL-32B-Thinking & 64 & 60.2 & 64.5 & 63.6 & 67.8 & 52.6 & 59.3 & \revsisecond{64.3} & 49.1 \\%
\midrule%
\rowcolor{revsilight}\multicolumn{11}{l}{\color{revsiink}\textbf{Our runs (Qwen3-VL-8B)}} \\%
Direct video\,\textsuperscript{O} & Qwen3-VL-8B & 32 & 47.3 & 41.5 & 49.8 & 70.5 & 39.8 & 52.9 & 39.9 & 36.8 \\%
Initial world\,\textsuperscript{O} & Qwen3-VL-8B & 32 & 42.5 & 40.0 & 40.9 & 41.6 & \revsibest{68.3} & 45.7 & 42.9 & 18.2 \\%
\rowcolor{revsiours}%
\textcolor{revsiink}{\textbf{Palimpsest\,\textsuperscript{O}}} & \textcolor{revsiink}{\textbf{Qwen3-VL-8B}} & \textcolor{revsiink}{\textbf{32}} & \revsibest{67.2} & 53.1 & 67.8 & 70.7 & \revsisecond{68.2} & \revsisecond{71.0} & \revsibest{83.9} & \revsisecond{55.8} \\%
\arrayrulecolor{revsiprimary}\bottomrule[1.1pt]\arrayrulecolor{black}%
\end{tabularx}%
\endgroup%
\arrayrulecolor{black}%

\caption{\textbf{ReVSI main results.} Dark and light shading mark the top two values per column. Avg.\ is the seven-task mean. Published scores use source-specific frame settings. Score sources are ReVSI (R)~\cite{revsi2026}, SpatialSpeak (S)~\cite{cao2026spatialspeak}, Video-ORA (V)~\cite{li2026videoora}, Render2Reason (T)~\cite{wu2026render}, GeoThinker-F (F)~\cite{li2026geothinkerf}, Cambrian-BEV (B)~\cite{siddiqui2026cambrianbev}, CROSS (C)~\cite{sun2026cross}, and our runs (O). Model references are VLM-3R~\cite{fan2025vlm3r}, GeoThinker~\cite{li2026geothinker}, SpatialStack~\cite{zhang2026spatialstack}, and Cambrian-S/P~\cite{yang2025cambrians,yang2026cambrianp}.}
\label{tab:types}
\end{table*}

\begin{figure*}[t]
\centering
\includegraphics[width=.83\textwidth]{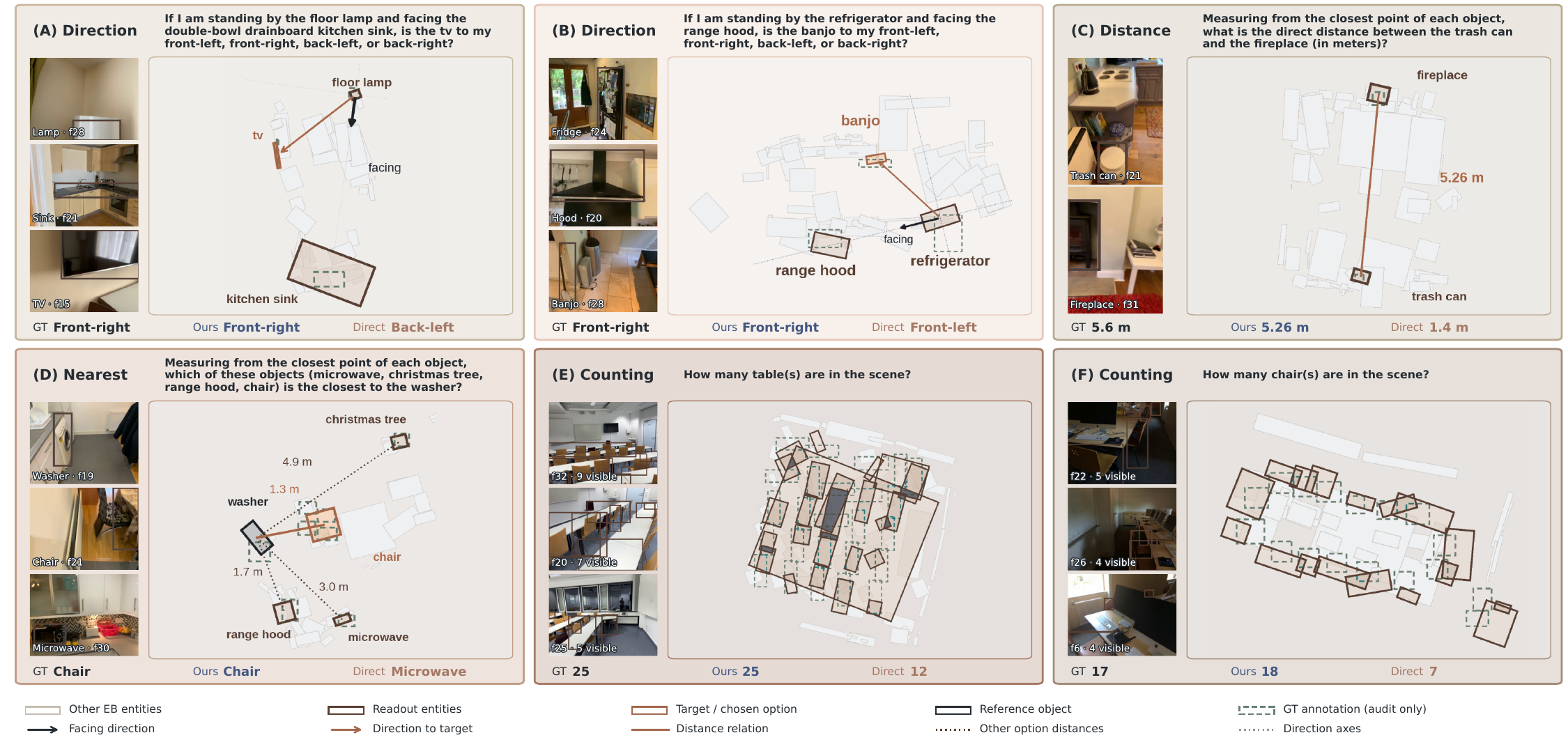}
\caption{\textbf{Selected ReVSI examples of spatial reasoning across views.}
Each panel shows sampled observations, a top view of the spatial world, the annotated answer (GT), and the answers from Palimpsest and direct video. A--B show direction, C absolute distance, D nearest-object selection, and E--F counting. Dashed green boxes show ground-truth annotations added for diagnosis. We selected cases with verified referents and better answers from Palimpsest.}
\label{fig:cases}
\end{figure*}

\noindent\begin{minipage}{\linewidth}
\begingroup
\setlength{\fboxsep}{4pt}%
\fcolorbox{palimpsestMocha!65}{palimpsestMocha!9}{%
\begin{minipage}{\dimexpr\linewidth-2\fboxsep-2\fboxrule\relax}
\footnotesize\raggedright
\textbf{Input} video $V$, trigger questions $Q_{\mathrm{trig}}$, and evaluation questions $Q_{\mathrm{eval}}$\par
\textcolor{palimpsestSubhead}{\textbf{1. Build}} Use metric depth and camera poses to lift tracked masks from $V$. Group observations into entities in $W$ and set their initial status.\par
\textcolor{palimpsestSubhead}{\textbf{2. Evolve}} Revisit views to check entity existence and names. Use $Q_{\mathrm{trig}}$ to identify unresolved references. Check video evidence, write accepted revisions to $W$, and record them in the ledger.\par
\textcolor{palimpsestSubhead}{\textbf{3. Read}} For each $q\in Q_{\mathrm{eval}}$, bind its referents in the revised $W$ and run its spatial program.\par
\textbf{Output} answers and revised world $W$.
\end{minipage}}%
\endgroup\par
\captionof{algorithm}{\textbf{Palimpsest inference.} The initial world is built from video before any question is seen.}
\label{alg:palimpsest}
\end{minipage}\par

\subsection{Execute queries over the revised world}

\Cref{alg:palimpsest} summarizes the build, revision, and readout steps. The reader resolves the referring phrases against the current world and passes the resulting entity records to a program for the question type. When the program needs a single referent, it ranks eligible matching entities by the number of supporting observations and breaks ties by their existence score. Accepted revisions can thus change either which entities are eligible for readout or which records a phrase retrieves.

In \cref{fig:method}, the question asks which object is closest to the microwave. The new binding lets the reader retrieve the existing box for \texttt{t11} as the geometry for \emph{multi-cooker}. Let $S(B)$ denote the sampled surface points of a stored box $B$. For a bound option $i$ and the reference object $r$, the reader computes
\begin{equation}
d_i=\min_{\mathbf x\in S(B_i),\,\mathbf y\in S(B_r)}
      \lVert\mathbf x-\mathbf y\rVert_2,
\qquad i^*=\arg\min_i d_i .
\end{equation}
Here $B_i$ and $B_r$ are the boxes retrieved for option $i$ and the reference object $r$. The Euclidean distance $d_i$ is in meters, and $i^*$ denotes the closest bound option. The binding changes which box enters this computation. The selected option is D, \emph{multi-cooker}.

Direction programs use relative positions, and counting programs group nearby entities and can use depth evidence to keep distinct objects separate. When the visual evidence for a count binding is exhausted, a majority rule can resolve it if the judgments favor one outcome. Size questions use a stored estimate for the queried phrase when available. This estimate comes from a direct model read of the video. Otherwise the reader uses an entity-level estimate from up to four views with the object marked, or falls back to the stored box extent. The reader stores these estimates apart from the boxes it uses to compute distances.

\palimpsestparagraph{Fusion with a spatial sketch.}
When a partial sketch is available, Palimpsest registers the video world to its coordinate frame and adds eligible video entities to the retained sketch instances.

\section{Experiments}

\subsection{Experimental setup}

ReVSI evaluates spatial reasoning from indoor videos and revises the scene annotations used in VSI-Bench~\cite{revsi2026}. It covers seven tasks, from counting objects and estimating sizes to judging distances and directions and planning routes. Reference answers depend on the sampled frames. For example, the correct object count can change when a frame subset excludes views of some instances. This protocol ties evaluation to the evidence available in the video and lets us test how different versions of the same world support answers to the same questions.

We evaluate on 6,158 questions across 380 scenes in the 32-frame ReVSI evaluation set. The direct video baseline answers from the benchmark's 32 frames with Qwen3-VL-8B~\cite{qwen3vl2025}, the same backbone used for Palimpsest's language-model calls. Palimpsest's perception tools process 64 frames. The comparison therefore evaluates complete systems with different inference costs. Palimpsest builds the initial world before receiving any questions and uses a public class vocabulary. In the video-only setting, neither system receives the benchmark's annotated geometry or answer labels.

For multiple-choice questions, we compare the first word of the response with the reference choice after stripping a trailing period and ignoring case. For a numerical prediction $\hat y$ and a positive reference value $y$, the per-question mean relative accuracy (MRA) is
\begin{equation}
\frac{1}{10}\sum_{k=0}^{9}
\mathbf{1}\!\left[\frac{|\hat y-y|}{y}\leq 0.50-0.05k\right].
\end{equation}
The indicator is one when the relative error meets the tolerance, which ranges from 50\% to 5\%. Missing or unparseable answers score zero. We average question scores within each benchmark subtype, then weight the subtypes equally within each reported task. The overall score averages the seven task scores and is reported on a 0--100 scale. We estimate 95\% intervals from 10,000 paired bootstrap resamples of scenes.

\subsection{Main results}

Palimpsest scores \textbf{67.2} on ReVSI~\cref{tab:types}. It improves over direct video answering by \textbf{19.9 points} with a 95\% confidence interval of [17.76, 22.05], and over readout from the initial world by 24.72 points [22.55, 26.98]. Direct video answering scores 47.32, while reading the initial world scores 42.49. The initial world already helps with room size, but its lower overall score shows how much accuracy the system loses when it reads the initial records directly. Averaging within each task group, final multiple-choice and numerical scores reach 70.24 and 64.95, respectively, compared with 43.19 and 50.41 for direct video. Final object-size performance remains close to that of direct video answering.

\subsection{Qualitative Studies}
\Cref{fig:cases} illustrates three ways the reader uses the revised world. The maps show the records available to the reader alongside dashed ground-truth boxes.

\emph{Viewpoint-relative direction (A--B).}
In A, the floor lamp sets the viewpoint and the sink sets the facing direction. The reader places the TV in the front-right quadrant. In B, the refrigerator and range hood define the viewpoint and facing direction, and the banjo is again front-right. In each case, the two reference entities define the question's coordinate frame in the reconstructed scene. The reader locates the target relative to that frame, so the result depends on the positions of all three entities. Both answers match the annotations. Direct video places the targets back-left in A and front-left in B.

\emph{Absolute and relative distance (C--D).}
In C, the reader measures the gap between the stored boxes for the trash can and fireplace. Its estimate is 5.26\,m against an annotated 5.6\,m, while direct video estimates 1.4\,m. In D, the same surface-distance calculation is applied to each option relative to the washer. It selects the chair, matching the annotation, while direct video selects the microwave. The two tasks place different demands on the geometry. C requires a metric estimate of the gap, while D requires the correct ordering of the option distances.

\emph{Counting across views (E--F).}
The reader counts entity instances in the scene world. In E, it finds 25 tables, matching the annotated count, while direct video reports 12. In F, it reports 18 chairs against an annotated count of 17. Direct video reports 7. The final world supports a substantially closer estimate in this example, with a residual error of one chair.

\subsection{Effects of world revision and readout}

\Cref{fig:ladder} reports gains over successive configurations. Evidence checks and observation acquisition add 6.27 points. Name verification and object recovery add 10.54 points. The final configuration combines remaining binding revisions with changes to counting, route parsing, and size estimation. It raises the score from 59.3 to 67.2, a gain of 7.91 points [6.49, 9.33], while scores decline on 489 questions.

In \cref{fig:method}, revisiting the video binds \emph{multi-cooker} to \texttt{t11}, previously labeled \emph{coffee maker}. Reading its existing box corrects the nearest-object answer. In a failure case, the agent binds a wall clock to a light.

\twocolumn[{%
\begin{minipage}{\textwidth}
\centering
\includegraphics[width=\textwidth]{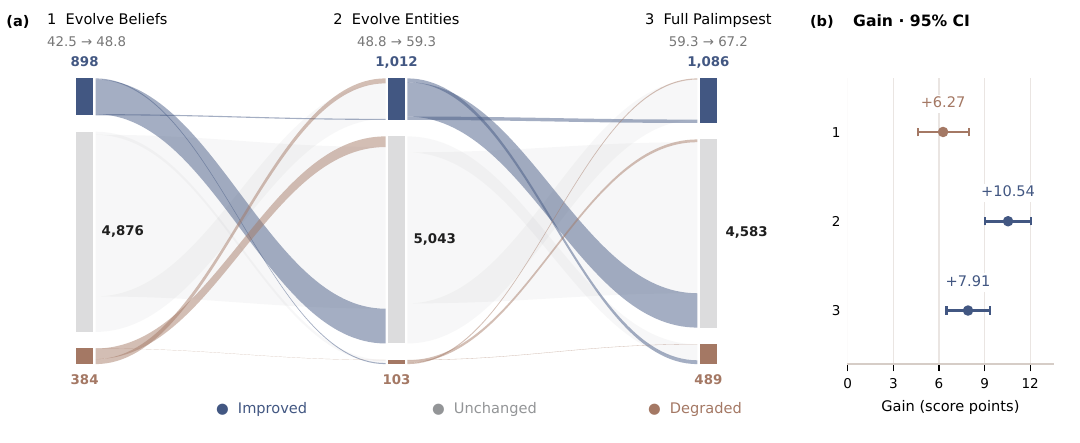}
\captionof{figure}{\textbf{Score improvement across three successive stages.}
Panel a groups the same 6,158 questions from 380 scenes by whether their scores improve, remain unchanged, or decline relative to the preceding configuration. Ribbons follow questions across these groups; width represents question count. The final stage, Full Palimpsest, combines binding revisions and readout corrections, with changes computed directly from the preceding Evolve Entities configuration.
Panel b shows seven-task macro-score gains with paired 95\% confidence intervals from 10,000 scene bootstrap resamples. All flows, gains, and intervals are recomputed from per-question scores; intervals are not added across the merged operations. Comparisons are order-dependent, and gains use unrounded scores.}
\label{fig:ladder}
\par\vspace{6pt}
\end{minipage}
}]
\enlargethispage{-6\baselineskip}
\subsection{Reuse of revisions across questions}
\label{sec:retention}

\begin{figure}[t]
\centering
\includegraphics[width=\linewidth]{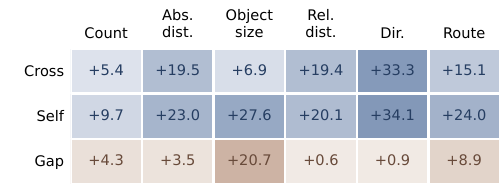}
\caption{\textbf{Reusing world revisions across questions.}
On all 380 ReVSI scenes, Cross and Self report gains over a world without question-triggered revisions. Triggers come from the other partition for Cross and the question's own partition for Self. Gap is Self minus Cross. Room size scores 68.2--68.3 across the conditions and is omitted.}
\label{fig:revision-reuse}
\end{figure}

We test whether revisions triggered by one group of questions help answer another group about the same scene. Across all 380 ReVSI scenes, we alternate which group can trigger revisions and answer all questions using the same binding and readout procedure. The score is 47.3 without question-triggered revision (None), 61.6 when the other group triggers revisions (Cross), and 67.1 when the question's own group triggers them (Self). Cross gains 14.2 points [12.2, 16.3], recovering 72\% of the Self gain. The Cross--Self gap is small for relative distance and direction and largest for object size~\cref{fig:revision-reuse}.

We test whether an object recovered for one question helps answer another question about the same scene. We partition each scene's questions by question-ID hash parity and vary which questions may trigger missing-object recovery. For each evaluated question, we compare three worlds: the world before recovery, the world revised using only questions in the opposite partition, and the world revised using all questions. We use the same reader in all three conditions. This comparison covers all 380 scenes and varies only the triggers for the missing-object recovery operator.

The score rises from 51.06 before recovery to 58.27 when only questions in the opposite partition trigger recovery, a gain of 7.21 points [5.89, 8.56]. Allowing all questions to trigger recovery raises the score to 59.30, an 8.25-point gain [6.89, 9.63]. Revisions triggered by other questions therefore recover 87\% of the gain available to this operator on the same benchmark questions.

Reuse depends on coverage of the recovery phrase. For 1,316 questions with a matching trigger in the other partition, scores rise from 14.0 to 54.8, compared with 55.1 under all-question recovery. Of these, 1,082 have a trigger from a different task type. For 165 questions requiring recovery without a matching trigger, scores remain at 4.7 and reach 39.3 under all-question recovery. The remaining 4,677 questions score 64.9 in every condition.

\twocolumn[{%
\begin{minipage}{\textwidth}
\centering
\includegraphics[width=\textwidth]{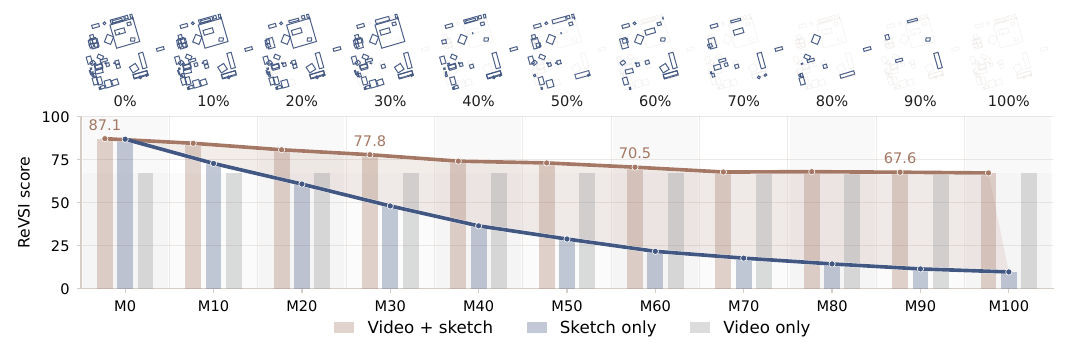}
\captionof{figure}{\textbf{Reasoning with partial spatial sketches on ReVSI.}
The top row shows sketches constructed from annotated object labels and 3D boxes. At each removal level we sample the omitted objects separately. Faded objects mark records withheld from the system. M0 supplies every record and M100 supplies none.
The bottom panel compares sketch-only, video-only, and fused readout over 380 scenes. The shaded band between the fusion and sketch-only curves visualizes the score gap. The gray reference marks the video-only score. At M100, no object records are provided, so fusion uses video alone.}
\label{fig:sketch}
\par\vspace{6pt}
\end{minipage}
}]
\enlargethispage{-7\baselineskip}
\subsection{Revision with partial spatial sketches}

A sketch is a list of object instances from the ReVSI annotations. Each supplied record contains a class label and an oriented 3D box. For each removal level $m$, we randomly withhold a fraction $m$ of the object records in each scene. We sample omissions separately at each level. The faded objects in \cref{fig:sketch} illustrate withheld records. At $m=0$ the sketch contains all records, and at $m=1$ it contains none.

Across the same 380 scenes, sketch-only reads from the retained boxes, video-only reads from the world built from video, and fusion uses both. When registration succeeds, fusion transforms the video entities into sketch coordinates and retains each supplied sketch box as a separate instance. A video entity is skipped if a retained sketch object shares its name and lies within the matching radius. Other video entities that pass the admission checks are added with their transformed geometry. When a query phrase exactly matches a retained sketch label, the reader binds it to that label. For counting, the resulting set can include both supplied instances and admitted video instances of the same class. If alignment fails, the reader keeps the coordinate frames separate. During counting, the reader preserves distinct sketch instances even when their boxes are close. Nearby video detections can still be grouped to avoid counting duplicate observations of one object.

As shown in \cref{fig:sketch}, fusion improves over the video-only world by 10.59 points when 30\% of the sketch records are removed and by 3.33 points [2.06, 4.68] at 60\% removal. With complete sketches, fusion scores only 0.30 points above sketch-only. Video can still introduce duplicate or spurious instances into a world that already contains the correct objects.

\section{Limitations and Conclusion}

Palimpsest builds an executable spatial world from video and revises its records against visual evidence. On 380 ReVSI scenes, it scores 67.2, exceeding direct video answering by 19.9 points. Both recovery-only and full-pipeline experiments show reuse by questions that did not trigger revision. Relative distance and direction retain most of their gains across partitions. The supplement also tests successive updates to one world on a fixed evaluation partition spanning 333 scenes. Later rounds improve its answers, although external question groups prescribe the triggers.

Object identity and extent remain limitations. The supplementary diagnostic finds higher annotated-object recall without an overall instance F1 gain. Semantic corrections can leave fragments and duplicates unresolved. Depth estimates lift visible surfaces, while occluded parts may lie outside the stored extent. With weak scale cues, vision-language models can use common-sense size priors and miss the dimensions of the actual instance. Palimpsest scores 70.7 on object size, close to 70.5 for direct video. Size also has the largest Cross--Self gap in the full-pipeline reuse test. Persistent revision improves access to evidence and records, but does not guarantee accurate physical measurements or autonomous convergence.

% Finish the body and flush any deferred figures before the end matter.
\clearpage
\section*{LLM Usage Statement}

Qwen3-VL was used as a component of the system and as an experimental baseline for video answering, vocabulary and name binding, visual judgments, and size estimation. Qwen3-VL-32B and Cambrian-P-7B were also used in separate research comparisons. Anthropic Claude assisted with debugging, monitoring ablation experiments, and the model-assisted revision audit. OpenAI Codex assisted with LaTeX formatting, page layout, and additional model-assisted revision audits.

{\footnotesize
\bibliographystyle{ieeenat_fullname}
\bibliography{references}
}

\clearpage
\setcounter{section}{0}
\setcounter{table}{0}
\setcounter{figure}{0}
\setcounter{equation}{0}
\setcounter{dbltopnumber}{2}
\setlength{\parfillskip}{0pt plus 1fil}
\renewcommand{\thetitle}{Palimpsest: Spatial Belief Improvement through Self-Evolving Agents}

\renewcommand{\theHsection}{supp.\arabic{section}}
\renewcommand{\theHtable}{supp.\arabic{table}}
\renewcommand{\theHfigure}{supp.\arabic{figure}}
\renewcommand{\theHequation}{supp.\arabic{equation}}
\maketitlesupplementary
\renewcommand{\thesection}{\Alph{section}}
\setcounter{section}{0}
\renewcommand{\thetable}{S\arabic{table}}
\renewcommand{\thefigure}{S\arabic{figure}}
\renewcommand{\theequation}{S\arabic{equation}}
\section{Implementation and Evaluation Details}
\label{supp:implementation}

Palimpsest constructs one executable world per video without benchmark questions or answer annotations. World inspection and question-triggered revision update the records read by spatial programs. We specify the interfaces, settings, and prompts used on all 380 scenes and 6,158 questions.

\subsection{Inputs and world records}

The language model uses 32 frames, a subset of the 64 frames processed by SAM~3 and DA3. Frame manifests preserve the decoded position, official source index, and image hash, linking each mask, camera, and revisit crop to its source image.

The world separates observations, entities, attributes, and geometry. An observation records its frame, mask reference, detector word, and confidence. An entity records its identifier, observation identifiers, label, support, and commitment status. Geometry is indexed by entity identifier and contains a metric center, oriented extent, and axes. Attributes contain candidate words and their observation counts, among other cached properties. Phrase bindings specify which names or entities a spatial program can access. A revision ledger records the changes emitted by each operation.

Question records contain the question and scene identifiers, task type, text, and options. The inference loader removes answer labels. Labels and 3D annotations enter only evaluation and post hoc audits, except in the separate partial-sketch setting, which explicitly supplies a specified subset of annotated object records.

\subsection{Vocabulary, observations, and geometry}

The vocabulary combines names proposed from video images with the public ScanNet200 category list. Vocabulary extraction uses the full image and four corner crops, each covering 60\% of image width and height. The image-listing prompt requests short singular object names without counts. Text-only filtering removes materials, parts, regions, and generic terms from the countable-object vocabulary. Wall, floor, and ceiling remain available as structure categories. Names and presence judgments are cached so that a repeated term does not require another identical judgment.

SAM~3 first checks category presence and then tracks retained categories through the perception frames. The initial fixed-vocabulary presence pass uses eight evenly spaced frames; the extension pass checks remaining categories over 32 selected positions. Tracking retains masks with at least 20 pixels and stores packed masks together with their frame and track identifiers. DA3-LARGE estimates depth and camera parameters, and DA3METRIC-LARGE supplies metric scale. Lifted mask points are represented in a shared, up-aligned metric coordinate frame.

\suppparagraph{Association and commitment.}
The builder initializes candidates from tracks. Different-word tracks can join when mask overlap reaches .8 often enough to cover at least half their combined frame lifetime. Same-word tracks can join when their centers are within .75\,m. Both joins respect a cannot-link test for spatially disjoint same-word masks in a shared frame. The builder also splits spatially separated portions of a track group and applies a .5\,m complete-linkage merge to resulting places that share a word. Complete linkage requires every cross-group center pair to satisfy that radius.

Cross-view point consensus retains points supported in at least half the other views in which they are surface-consistent. Surface agreement allows a depth difference of at most the larger of .3\,m and 15\% of the observed depth. The entity center is estimated using the adopted median-based 3D rule, and its extent and axes are fitted to the retained points. If $O_e$ is the set of observed frames and $V_e$ the expected visible frames for entity $e$, initial coverage is $|O_e|/|O_e\cup V_e|$. An entity is committed when it has at least two observed frames or coverage at least .5. A minor place split from a group requires at least three observed frames. Uncommitted hypotheses retain their records for subsequent inspection.

\begin{table}[!htbp]
\centering
\caption{\textbf{Adopted construction and revision settings.} The builder and final count program use different grouping operations. The two inspection sweeps apply the same per-scene budget to their respective input worlds. A sweep can stop before its limits when no eligible evidence check remains.}
\label{tab:supp-settings}
\small
\setlength{\tabcolsep}{4pt}
\begin{tabularx}{\columnwidth}{Xr}
\toprule
\rowcolor{suppHeader}
Setting & Value \\
\midrule
Cross-word mask overlap & .8 \\
Same-word track association radius & .75\,m \\
Place merge radius, complete linkage & .5\,m \\
Judgments per scene per inspection sweep & 1,000 maximum \\
Rounds per inspection sweep & 6 maximum \\
Phrase verification views & 5 maximum \\
Recovery detection confidence threshold & .05 \\
Alias proposal names / candidate views & 3 / 8 maximum \\
Final count grouping radius & .75\,m \\
\bottomrule
\end{tabularx}
\end{table}

\subsection{Evidence selection and revision}

Inspection prioritizes unresolved existence and name checks with weak support or conflicting words, then considers other open checks. Evidence comes from the entity's observed masks in the allowed language-model frames. The first crop favors a large observation; later crops favor a different viewing direction and visible area. A frame is not reused for the same check.

For each word-level check, the system accumulates yes and no responses until one leads by two. It removes rejected words and can withdraw an entity when all its eligible words have been rejected. A proposed name opens a new check; only a confirmed proposal can rename an entity or restore a hypothesis to committed status. Each round uses the updated entity state and judgment history. Resolved checks leave the open pool, while new name proposals enter it. Checks with no remaining usable evidence stay unresolved. A sweep ends after six rounds, 1,000 judgments per scene, or exhaustion of eligible checks. After acquiring additional observations, the system runs a second sweep under the same rules. Our experiments assess the revised records and their reuse, without separately measuring the benefit of adapting inspection choices between rounds.

Question-triggered stages first identify unresolved phrases and acquire observations for their candidate names. Bound entities that lack the exact phrase can undergo phrase-specific visual checks. These checks use the same two-vote margin and at most five views, with projected-view fallback enabled for entities with insufficient observed views. Missing-object recovery applies SAM~3 image detection at confidence .05 and records the resulting entity or promotion. This stage uses the question's presupposition that its named object is present; the presupposition guides a search and is not an answer annotation.

The final alias-binding stage considers a phrase that remains unresolved. A text prompt selects up to three existing scene names, after which the system collects up to eight candidate entity views. With multiple candidates, the model selects an entity from marked views. A single available candidate is selected without an additional visual comparison. The update adds the phrase to the entity's words, commits the selected hypothesis when necessary, and records the candidate selection in the ledger. Existing centers and extents are retained by this binding operation. The updated name hash is used to keep the binding cache consistent with the revised world.

\subsection{Spatial programs}

The final reader accesses the revised world and its associated caches. It does not generate a free-form answer from a rendered map. Counting groups the entities bound to each phrase, using the adopted .75\,m rule and recorded distinct-instance or negative-evidence checks. Object size prefers a stored phrase-specific estimate, then the entity-level longest-side estimate, and finally the geometric extent. These scalar estimates remain distinct from the boxes used for spatial distance calculations.

Absolute distance uses the minimum distance between sampled surfaces of the selected oriented boxes. Relative-distance tasks compute the same quantity between the reference and each option and select the closest or farthest. Direction uses the horizontal vector from the standing position to the facing object as the reference heading, reversing it when the question asks for the opposite direction. The easy direction labels use counterclockwise boundaries at 135 and 225 degrees; the hard labels use four 90-degree sectors. Room-size tasks read the stored room-area estimate and combine rooms as required by the question.

Route planning parses the start, initial facing object, intermediate waypoints, and goal. The program computes successive horizontal headings and turns from their entity centers. Signed angles from 45 to 135 degrees indicate left turns; the corresponding negative interval indicates right turns; magnitudes of at least 135 degrees indicate turning back. The reader normalizes the instructional note before locating action blanks, then selects the option agreeing with the computable turns. This is a waypoint-based geometric readout, rather than collision-aware navigation through a reconstructed free-space map.

\subsection{Model configuration and reproducibility}

The main language model is Qwen3-VL-8B-Instruct, served through the vLLM OpenAI-compatible API. The recorded main-run server log identifies vLLM 0.31.0, bfloat16 weights, tensor parallelism of one, a context limit of 32,768 tokens, and at most 32 images per request. Requests use temperature zero. The execution machine has eight RTX~5880 Ada GPUs; the drivers use available devices for perception and independent model services. Separate environments support geometry, SAM~3, and language-model serving.

The reproduction files provide scene lists, prompts, parameters, source hashes, and stage-to-implementation mappings. Inspected sources match the archived main-run hashes. Package versions describe the deployment at verification time; the historical server log separately records the main-run vLLM configuration.

\begin{table*}[!htbp]
\centering
\caption{\textbf{Deployment environments available at verification.} The geometry, SAM~3, and language-model services use separate Python environments. These versions describe the inspected deployment; the recorded main-run log independently identifies vLLM 0.31.0 and the adopted serving arguments. Source-file hashes are checked against the archived run manifest.}
\label{tab:supp-software}
\small
\setlength{\tabcolsep}{7pt}
\begin{tabular}{lllll}
\toprule
\rowcolor{suppHeader}
Environment & Python & PyTorch & Transformers & vLLM \\
\midrule
Geometry & 3.10.19 & 2.8.0+cu128 & 4.57.1 & --- \\
SAM~3 & 3.11.17 & 2.5.1+cu121 & 4.57.1 & --- \\
Language-model service & 3.12.3 & 2.13.0+cu129 & 5.17.0 & 0.31.0+cu129 \\
\bottomrule
\end{tabular}
\end{table*}

Saved-state replay evaluates archived worlds and caches, including the factorial comparisons in \cref{supp:world-readout}. Fresh construction also requires model weights, videos, perception outputs or tool runtimes, and vocabulary judgments. The code package provides construction and reading entry points; the reproduction index lists the stage drivers and saved inputs for each analysis.

\subsection{Evaluation and artifact index}

Multiple-choice questions use exact answer matching. For a numerical question with positive annotated value $y_i$ and parsed answer $\widehat y_i$, the question score is
\begin{equation}
a_i=\frac{1}{10}\sum_{j=0}^{9}\mathbf{1}\!\left[
\frac{|\widehat y_i-y_i|}{y_i}\leq .50-.05j\right].
\end{equation}
Missing or unparseable answers receive zero. We average question scores within each benchmark subtype, weight the subtypes equally within each reported task, and average the seven task scores for the overall result on a 0--100 scale. Counts of improved or declined questions compare individual scores; they do not replace this task-macro aggregation.

The artifact index links each experiment to its scene list, code, outputs, and scorer. Audit instructions and per-event judgments accompany the revision audit. GT-assisted audit and scoring code remain separate from inference. Each analysis specifies any changes to question sets, resampling counts, or denominators.

\subsection{Prompt templates}

The following templates are copied from the implementation. Braced fields are replaced by scene names, query phrases, or candidate counts at runtime. Image-based prompts are accompanied by the corresponding crops, marked candidate views, or scene frames. Model responses are parsed into the expected list, yes/no judgment, candidate index, or scalar; the templates are not rewritten for individual evaluation answers.

P11 uses a positional placeholder, \texttt{\{\}}, which is filled with the object phrase at runtime. The stored size is a VLM estimate, separate from the geometric box extent.

% Generated verbatim from implementation constants; see reproduction/prompts.json.
\begin{suppprompt}{P1. Scene vocabulary}
List every distinct kind of physical object you can see in this image of an indoor scene (furniture, appliances, fixtures, decor, small items). Use short singular lowercase English nouns, one entry per kind, no counts, no adjectives. Answer only with a JSON list of strings, at most 40 entries.
\end{suppprompt}
\begin{suppprompt}{P2. Whole-object vocabulary filter}
Is "\{w\}" the name of a whole physical object that one could count in a room (for example furniture, an appliance, a device, a container, a plant, a decoration)? Answer no for a material (metal, plastic), a part of an object (leg, handle, hinge, rim), a surface or region (surface, edge, corner, shadow, reflection, floor tile), a generic word (object, item, thing), or text. Answer yes or no only.
\end{suppprompt}
\begin{suppprompt}{P3. Existence and word verification}
Look at the object in the middle of this image. Is it a \{w\}? Answer yes or no.
\end{suppprompt}
\begin{suppprompt}{P4. Entity naming}
What is the object in the middle of this image? Answer with its common name in one to three words only.
\end{suppprompt}
\begin{suppprompt}{P5. Generic-name filtering}
Is "\{w\}" a general category word (such as appliance, device, fixture, furniture, equipment, item, object, container) rather than the name of a specific kind of object? Answer yes or no.
\end{suppprompt}
\begin{suppprompt}{P6. Initial phrase binding}
Objects detected in a scene, by name: \{names\}.\par Which of these names refer to the same kind of object as "\{phrase\}"? Include synonyms and more specific or more general names for the same object; exclude different objects and parts of objects. Answer only with a JSON list of names copied from the list above, or [] if none.
\end{suppprompt}
\begin{suppprompt}{P7. Semantic candidate verification}
In a room, someone asks about "\{phrase\}". Does the object name "\{name\}" refer to a \{phrase\} itself (the same object category, a synonym, or a more specific kind of it)? Answer no if it is only a part of a \{phrase\}, something attached to, placed on or used with a \{phrase\}, or a different kind of object. Answer with yes or no only.
\end{suppprompt}
\begin{suppprompt}{P8. Alias name proposals}
A \{p\} is in this scene, but an object detector may have called it by another name. Names the detector used in this scene:\par \{names\}\par Which of these names could be the \{p\}? Answer only with a JSON list of up to 3 names copied from the list, best first.
\end{suppprompt}
\begin{suppprompt}{P9. Alias candidate selection}
These \{k\} images are numbered 1 to \{k\} in order. Each marks one object with a green box. The scene contains a \{p\}. Which image marks the \{p\}? Answer with the image number only.
\end{suppprompt}
\begin{suppprompt}{P10. Entity-level size estimate}
These \{n\} images come from one video of a room. The green box in each image marks the same object, a \{w\}. Estimate the length of its longest dimension (length, width, or height) in centimeters, using the room around it for scale. Answer with a number only.
\end{suppprompt}
\begin{suppprompt}{P11. Phrase-level size estimate}
Based on visual evidence from the video, what is the length of the longest dimension (length, width, or height) of the \{\}, measured in centimeters?
\end{suppprompt}

\section{World Revision and Readout}
\label{supp:world-readout}

\subsection{Separating world and reader changes}
The main configurations progress from 42.5 to 48.8 and then 59.3, gaining 16.8 points through inspection, additional observations, and recovery. Our factorial analysis starts from this revised state and examines the final 59.3--67.2 change. We replay earlier and final records ($W_0$, $W_1$) with earlier and final readers ($R_0$, $R_1$) on 380 scenes and 6,158 questions. Records include entities, attributes, bindings, and evidence caches. Both original combinations reproduce every saved answer without new model calls. Evidence acquisition differed when the records were produced. Intervals use 10,000 paired scene resamples with seed 20261106; gains use unrounded scores.

\begin{table*}[!htbp]
\centering
\caption{\textbf{World revision and readout in a factorial comparison.} This final-stage replay uses the same 380 scenes and 6,158 questions; earlier revision stages are already included in both worlds. $W_0$ and $W_1$ contain the earlier and final saved world records and their associated caches; $R_0$ and $R_1$ are the corresponding readers. The original combinations reproduce every saved answer. Count, Abs., Size, Room, Rel., Dir., and Route denote the seven ReVSI tasks.}
\label{tab:supp-factorial}
\small
\setlength{\tabcolsep}{7pt}
\begin{tabular}{llrrrrrrrr}
\toprule
\rowcolor{suppHeader}
World & Reader & Count & Abs. & Size & Room & Rel. & Dir. & Route & Overall \\
\midrule
$W_0$ & $R_0$ & 49.8 & 65.7 & 51.6 & 68.2 & 68.8 & 79.7 & 31.4 & 59.3 \\
$W_0$ & $R_1$ & 49.8 & 65.7 & 51.6 & 68.2 & 68.8 & 79.7 & 55.4 & 62.7 \\
$W_1$ & $R_0$ & 49.8 & 67.8 & 52.6 & 68.2 & 71.0 & 83.9 & 32.2 & 60.8 \\
\rowcolor{suppHighlight}
$W_1$ & $R_1$ & 53.1 & 67.8 & 70.7 & 68.2 & 71.0 & 83.9 & 55.8 & 67.2 \\
\bottomrule
\end{tabular}
\end{table*}

\begin{table}[!htbp]
\centering
\caption{\textbf{Conditional effects and interaction.} Gains are overall score points with paired 95\% scene bootstrap intervals. Improved and declined count questions whose individual scores increase or decrease. The interaction is the difference between the reader effects on $W_1$ and $W_0$.}
\label{tab:supp-factorial-effects}
\small
\setlength{\tabcolsep}{3pt}
\begin{tabular}{lrrr}
\toprule
\rowcolor{suppHeader}
Comparison & Gain [95\% CI] & Improved & Declined \\
\midrule
World, $R_0$ fixed & +1.5 [0.8, 2.2] & 136 & 9 \\
World, $R_1$ fixed & +4.5 [3.7, 5.3] & 997 & 462 \\
Reader, $W_0$ fixed & +3.4 [2.3, 4.6] & 89 & 27 \\
Reader, $W_1$ fixed & +6.4 [5.2, 7.7] & 961 & 489 \\
Interaction & +3.0 [2.4, 3.6] & --- & --- \\
\rowcolor{suppHighlight}
Combined change & +7.9 [6.5, 9.3] & 1,086 & 489 \\
\bottomrule
\end{tabular}
\end{table}

The final world adds alias bindings, VLM size records, and counting-verdict caches. Their benefit depends on which fields the reader accesses. With $R_1$ fixed, changing $W_0$ to $W_1$ adds 4.5 points [3.7, 5.3]. The older $R_0$ does not read the new phrase-size field or final counting caches; its corresponding gain is 1.5 points [.8, 2.2]. Distance and direction improve under either reader, consistent with better bindings exposing existing geometry. This comparison does not isolate alias binding from every other record change. Reader changes add 3.4 points on $W_0$ and 6.4 on $W_1$; the interaction is 3.0 [2.4, 3.6]. Together, the changes improve 1,086 questions and lower scores on 489.

Route increases from 31.4 to 55.4 on unchanged $W_0$. The old parser treated a placeholder inside the explanatory note as an action blank, shifting option alignment. Removing the note before parsing lets the reader match computed turns to the intended slots. This is a readout correction, not evidence that the world geometry changed. Size similarly depends on the interface: $R_1$ prefers a cached phrase-level VLM estimate when available, whereas $R_0$ follows the earlier entity-level path. On $W_1$, size rises from 52.6 to 70.7. The estimate supplements incomplete geometric evidence but is not restricted to cases with a diagnosed occlusion, and it does not refine the reconstructed box. These dependencies explain why new records and a compatible reader must be evaluated together.

\subsection{Vocabulary and evidence controls}
\label{supp:vocabulary-control}
Under the Cross protocol, only the opposite question partition triggers revisions. Track adds SAM~3 tracks for trigger phrases; Detect also promotes or detects unresolved referents; Write further adds size records and final binding. Write omits post-reconstruction existence/name inspection, phrase verification, pairwise identity checks, and remaining-reference verdicts. Its binding step still uses candidate views. All arms use the final reader without omitted verdict caches. None retains front-end inspection, so these are reduced-pipeline controls, not isolated single-operator ablations.

\begin{table}[!htbp]
\centering
\caption{\textbf{Vocabulary and writing controls in the Cross protocol.} The final reader answers the same 6,158 questions in every condition. Track, Detect, and Write use the opposite partition as triggers; None and Full Cross are the saved parity-protocol conditions. Empty counts unanswered questions. Scores are task-macro means, with gains computed before rounding.}
\label{tab:supp-vocabulary-controls}
\small
\setlength{\tabcolsep}{5pt}
\begin{tabular}{lrr}
\toprule
\rowcolor{suppHeader}
Condition & Overall & Empty \\
\midrule
None & 47.3 & 1,339 \\
Track & 47.0 & 1,079 \\
Detect & 54.4 & 334 \\
Write & 55.7 & 191 \\
\rowcolor{suppHighlight}
Full Cross & 61.6 & 197 \\
\bottomrule
\end{tabular}
\end{table}

Track differs from None by $-.3$ points [$-2.2$, 1.6]. Detect gains 7.1 points [4.8, 9.4] over None; Write adds 1.3 [.6, 2.0]. Full Cross remains 5.9 points [4.6, 7.3] above Write and 7.2 [5.7, 8.7] above Detect. Write has slightly fewer empty answers than Full Cross, so returning more answers alone does not explain their score difference. The contrast measures the omitted operations jointly, without matching total inference cost.

On 645 registered scene--partition entries from 324 scenes, Full Cross versus Write increases recall by .9 percentage points [.3, 1.6], lowers precision by 2.1 [$-2.9$, $-1.3$], and raises duplicates by 10.3 [9.3, 11.3]. Of 760 entries, 115 fail registration. Metrics follow \cref{supp:world-quality}; better answers do not imply a cleaner instance inventory.

\subsection{Frame controls}
\label{supp:frame-controls}
We first give the direct reader all 64 perception frames while keeping the 32-frame question set fixed at 380 scenes and 6,158 questions. Its gain over 32-frame direct video is 2.0 points [1.1, 2.8]; Palimpsest remains 17.9 points [15.7, 20.2] higher. We also rerun the complete pipeline on the previously fixed 99 scenes and 1,645 questions using only the official 32 frames for language and perception. Depth, scale, SAM~3 processing, reconstruction, revisits, and recovery are recomputed without 64-frame perception caches.

\begin{table}[!htbp]
\centering
\caption{\textbf{Frame controls on a fixed question set.} The full-set comparison covers 380 scenes and 6,158 questions. Every row in the perception-rerun block uses the same 99 scenes and 1,645 questions. VLM / tools gives the language-model and perception-frame counts; direct readers have no separate tool input. Comparisons are made within each block.}
\label{tab:supp-frame-controls}
\small
\setlength{\tabcolsep}{4pt}
\begin{tabular}{llr}
\toprule
\rowcolor{suppHeader}
Condition & VLM / tools & Overall \\
\midrule
\multicolumn{3}{l}{\emph{380 scenes}} \\
Direct & 32 / --- & 47.3 \\
Direct & 64 / --- & 49.3 \\
Palimpsest & 32 / 64 & 67.2 \\
\midrule
\multicolumn{3}{l}{\emph{99-scene perception rerun}} \\
Direct & 32 / --- & 48.8 \\
Direct & 64 / --- & 49.9 \\
Palimpsest & 32 / 64 & 67.2 \\
\rowcolor{suppHighlight}
Palimpsest & 32 / 32 & 65.5 \\
\bottomrule
\end{tabular}
\end{table}

The 32-frame pipeline differs from the 64-frame perception setting by $-1.7$ points [$-5.2$, 1.3]. It exceeds 32-frame direct video by 16.7 points [12.3, 21.1] and 64-frame direct video by 15.6 [10.9, 20.4]. The system benefit persists with matched frame access. Compute remains unmatched, and these controls do not identify how much of the main gain comes from additional frames.

\section{Revision Reliability and World Quality}
\label{supp:revision-audit}

\subsection{Evidence audit of accepted edits}
We sample 30 events without replacement from each of eight operation types across the 380 final worlds, using seed 20261118. Claude reviews the recorded change, up to four marked views, and nearby annotations when registration is available, without QA scores. Judgments are correct, incorrect, or undetermined. Annotations are used only after inference. This is a model-assisted audit, not human validation; the exact judge revision was not recorded.

The unit is one accepted edit, not a complete entity or an answer. The marked object must support its name or status change; insufficient evidence is undetermined. For word removal, the eligible population is the 4,611 updates that remove at least one word, excluding additive-only changes. The final sample contains 30 actual removals and 240 eligible judgments overall.\footnote{Eligibility screening replaced 19 additive-only draws with 19 actual-removal draws, without replacement, using seed 20261119. The original and replacement draws remain in the audit manifest.}

\begin{table*}[!htbp]
\centering
\caption{\textbf{Evidence audit of accepted revisions.} Population counts refer to eligible ledger events across 380 scenes. Thirty eligible events are audited per operation. C, I, and U denote correct, incorrect, and undetermined model-assisted judgments. Precision is $C/(C+I)$ with a Wilson 95\% interval. The last column treats U as incorrect. The aggregate weights each operation by its population count; its margins are 1.96 times the stratified standard error. Word removal uses the 4,611 actual-removal updates; eligibility screening is documented in the sampling protocol.}
\label{tab:supp-revision-audit}
\small
\setlength{\tabcolsep}{6pt}
\begin{tabular}{lrrrrrrr}
\toprule
\rowcolor{suppHeader}
Operation & Population & Audited & C & I & U & Precision [95\% CI] & U as incorrect \\
\midrule
Entity withdrawal & 3,018 & 30 & 24 & 2 & 4 & .92 [.76, .98] & .80 \\
Entity promotion & 182 & 30 & 22 & 3 & 5 & .88 [.70, .96] & .73 \\
Relabeling & 4,011 & 30 & 24 & 2 & 4 & .92 [.76, .98] & .80 \\
Word removal & 4,611 & 30 & 26 & 1 & 3 & .96 [.82, .99] & .87 \\
Word addition & 6,893 & 30 & 21 & 6 & 3 & .78 [.59, .89] & .70 \\
Recovery detection & 222 & 30 & 23 & 5 & 2 & .82 [.64, .92] & .77 \\
Recovery promotion & 38 & 30 & 12 & 15 & 3 & .44 [.28, .63] & .40 \\
Phrase binding & 43 & 30 & 16 & 10 & 4 & .62 [.43, .78] & .53 \\
\midrule
\rowcolor{suppHighlight}
Population-weighted estimate & 19,018 & 240 & 168 & 44 & 28 & $.877\pm.065$ & $.777\pm.076$ \\
\bottomrule
\end{tabular}
\end{table*}

The weighted estimate is .877 with margin .065 among determined judgments and .777 with margin .076 when undetermined cases count as incorrect. It describes edit events, not complete world entities. Recovery promotion and final binding have lower precision, .44 and .62. These fallbacks account for 38 and 43 events, respectively: 81 of 19,018 eligible edits, or 0.43\%. Promotion uses an exact-name hypothesis when no committed referent is available; final binding handles phrases still unresolved later. Promotion can precede a new detection, so it is not conditioned on every detector having failed. Their small event share does not imply negligible QA impact, since one edit can serve several questions. Attempts and rejection counts are incomplete, so these are not acceptance rates.

\subsection{Object coverage and instance quality}
\label{supp:world-quality}
We compare initial construction (T), question-free inspection (pn), question-triggered observations and recovery (P), and final records (NEW). Initial worlds are available for 335 scenes. The initial-world registration fails the four-inlier diagnostic requirement in 54, leaving 281 scenes and 3,846 visible annotated objects. These exclusions concern post hoc GT alignment, not inference failures or missing outputs in the 380-scene QA evaluation. Every stage uses the same registration fitted on T and one name-binding cache over all stages. A center inside an oriented annotation box expanded by .3\,m matches that instance; further matches count as duplicates.

Precision divides represented instances by class-bound entity records; recall divides them by visible annotated instances. Duplicate, other-class, and no-box rates use the class-bound denominator. Multiple class bindings can include the same entity; annotation-unbound entities are excluded and reported separately. This tests centers and class bindings, not full box or surface accuracy.

\begin{table*}[!htbp]
\centering
\caption{\textbf{World quality on 281 consistently registered scenes.} All stages share the initial-world registration and class-binding cache. Precision (Prec.), recall (Rec.), F1, duplicate (Dup.), other-class (Other), and no-box rates are percentages for visible annotated classes. The last two columns give committed and annotation-unbound entity counts per scene.}
\label{tab:supp-world-quality}
\small
\setlength{\tabcolsep}{6pt}
\begin{tabular}{lrrrrrrrr}
\toprule
\rowcolor{suppHeader}
Stage & Prec. & Rec. & F1 & Dup. & Other & No box & Committed & Unbound \\
\midrule
T & 38.1 & 74.1 & 50.4 & 19.1 & 25.4 & 17.4 & 114.0 & 90.2 \\
pn & 35.7 & 74.9 & 48.3 & 30.1 & 18.8 & 15.5 & 106.4 & 81.0 \\
P & 36.1 & 78.4 & 49.5 & 29.3 & 19.1 & 15.4 & 107.5 & 81.2 \\
\rowcolor{suppHighlight}
NEW & 36.2 & 78.8 & 49.6 & 29.2 & 19.1 & 15.4 & 107.5 & 81.2 \\
\bottomrule
\end{tabular}
\end{table*}

Committed entities per scene fall from 114.0 to 107.5, and annotation-unbound entities from 90.2 to 81.2. Other-class and no-box rates fall from 25.4\% to 19.1\% and from 17.4\% to 15.4\%, respectively. Unbound entities may still exist; no-box rate does not measure missed-object recall. These reductions do not imply that every entity improves.

Name changes also affect how fragments enter the diagnostic. Previously unbound fragments can enter the class-bound denominator; already bound records can move from Other to Dup when their names are corrected. The rise in duplicates during pn is consistent with this reclassification mechanism, but the aggregate table does not quantify its share. Duplicate entries remain unresolved instances under the metric, rather than a purely cosmetic denominator effect.

From pn to NEW, recall rises from 74.9\% to 78.8\%, precision from 35.7\% to 36.2\%, and F1 from 48.3\% to 49.6\%; duplicates decline from 30.1\% to 29.2\%. Across the complete T--NEW comparison, recall gains 4.7 points [3.8, 5.6], precision falls 2.0 [$-2.9$, $-1.0$], and F1 changes by $-.8$ [$-1.7$, .2]. Duplicates increase by 10.1 points [9.0, 11.3], with median matched-center error .264\,m versus .268\,m. Later question-triggered stages therefore improve on pn, while the full trajectory still leaves instance precision and geometry unresolved.

\subsection{Human verification of later-round edits}
\label{supp:continuous-audit}
For the sequential experiment in \cref{supp:continuous-revision}, we verify the inherited ledger prefix and sample only new events that actually change entities or words. S1--S2 contains 76 eligible semantic/status events and S2--S3 contains 51. We draw 20 per transition uniformly without replacement with seed 20261127. Each event retains before/after records and saved detector or binding-candidate views; projected-box fallbacks are tagged. The initial Codex-assisted review checks visual support without QA outcomes and records a reason. One human reviewer then checks the same evidence under the same criteria, without access to QA outcomes, and confirms the C/I/U judgments. We report single-reviewer human verification of the assisted review, not a separately blinded annotation study. Scalar estimates and auxiliary-cache verdicts are excluded.

\begin{table*}[!htbp]
\centering
\caption{\textbf{Human-verified later semantic and status revisions.} Twenty events per transition are sampled uniformly before review. The assisted C/I/U judgments are unchanged after human verification. Pop.\ gives the eligible population; C, I, and U denote correct, incorrect, and undetermined judgments. Precision excludes U, while the final column counts it as incorrect. Both rates have descriptive Wilson 95\% intervals and assess the edit rather than geometric accuracy or track purity.}
\label{tab:supp-continuous-audit}
\small
\setlength{\tabcolsep}{7pt}
\begin{tabular}{lrrrrrrr}
\toprule
\rowcolor{suppHeader}
Transition & Pop. & Sampled & C & I & U & Precision [95\% CI] & U as incorrect [95\% CI] \\
\midrule
S1--S2 & 76 & 20 & 15 & 2 & 3 & .882 [.657, .967] & .750 [.531, .888] \\
S2--S3 & 51 & 20 & 17 & 3 & 0 & .850 [.640, .948] & .850 [.640, .948] \\
\bottomrule
\end{tabular}
\end{table*}

The sample contains 34 detections, five bindings, and one promotion. Correct existence or naming can coexist with off-target observations, duplicates, and inaccurate geometry. These small samples do not establish a quality difference between rounds; their unequal population sizes also preclude interpreting an unweighted pooled rate as population correctness. The audit checks individual edits, while the score experiment measures their collective utility.

\section{Persistence and Reuse Across Questions}
\label{supp:revision-reuse}

\subsection{Cross question evaluation}
Questions are split by MD5 identifier parity within each scene. Starting from the same saved front end, each partition triggers revisions separately. None reads the front end without question-triggered revision; Cross uses the opposite partition's revisions; Self uses the evaluated question's own partition. All questions use the same binding and final reader. Trigger restrictions do not restrict which questions can be answered. The front-end inspection is shared, and initial construction sees no questions.

\begin{table*}[!htbp]
\centering
\caption{\textbf{Reuse across question partitions on all 380 scenes.} None excludes question-triggered revision, Cross uses only the opposite partition as revision triggers, and Self uses the question's own partition. Each row uses the same binding and reader across its three conditions. The recovery-only experiment isolates missing-object recovery with the earlier reader; the full-pipeline experiment uses the final reader and all question-triggered stages. Their baselines therefore differ. The reuse share is the Cross gain divided by the Self or Full gain. Self denotes own-partition triggers for the full pipeline; Full denotes all-question triggers for recovery only.}
\label{tab:supp-reuse-overall}
\small
\setlength{\tabcolsep}{6pt}
\begin{tabular}{lrrrrrr}
\toprule
\rowcolor{suppHeader}
Revision setting & None & Cross & Self / Full & Cross $-$ None [95\% CI] & Reference gain [95\% CI] & Reuse share \\
\midrule
Missing-object recovery & 51.1 & 58.3 & 59.3 & +7.2 [5.9, 8.6] & +8.3 [6.9, 9.6] & 87\% \\
\rowcolor{suppHighlight}
Full pipeline & 47.3 & 61.6 & 67.1 & +14.2 [12.2, 16.3] & +19.8 [17.8, 21.9] & 72\% \\
\bottomrule
\end{tabular}
\end{table*}

On 380 scenes and 6,158 questions, Cross gains 14.2 points [12.2, 16.3], retaining 72\% of Self's gain. It improves 1,388 questions, lowers scores on 247, and leaves 4,523 unchanged. Partitioned Self scores 67.1 versus 67.2 for the saved all-question main run; their difference is .1 points [$-.2$, .4]. Missing-object recovery alone, with its earlier reader fixed, gives a separate gain of 7.2 points [5.9, 8.6], or 87\% of all-question recovery's gain. It improves 625 questions and lowers scores on 14. The two baselines differ in their records and readers. Overall intervals use 10,000 paired scene resamples with seed 20261106.

\begin{table}[!htbp]
\centering
\caption{\textbf{Task scores under the full-pipeline reuse protocol.} The same final reader answers every condition. Cross excludes all evaluation-partition questions from revision triggers, while Self includes them. Relative distance and direction retain most of the Self benefit; object size retains less. Overall is the seven-task mean.}
\label{tab:supp-reuse-tasks}
\small
\setlength{\tabcolsep}{7pt}
\begin{tabular}{lrrr}
\toprule
\rowcolor{suppHeader}
Task & None & Cross & Self \\
\midrule
Counting & 43.2 & 48.5 & 52.9 \\
Absolute distance & 44.9 & 64.3 & 67.8 \\
Object size & 42.9 & 49.8 & 70.5 \\
Room size & 68.3 & 68.3 & 68.2 \\
Relative distance & 50.7 & 70.2 & 70.8 \\
Direction & 49.7 & 82.9 & 83.8 \\
Route planning & 31.8 & 46.9 & 55.8 \\
\midrule
\rowcolor{suppHighlight}
Overall & 47.3 & 61.6 & 67.1 \\
\bottomrule
\end{tabular}
\end{table}

\subsection{Coverage of reusable referents}
Of all questions, 1,316 require recovery and have a matching trigger phrase in the other partition. Their question-mean score rises from 14.0 to 54.8, close to 55.1 under all-question recovery; 1,082 have a trigger from a different task. Another 165 require recovery without a matching trigger: they remain at 4.7 in Cross and reach 39.3 under all-question recovery. The remaining 4,677 stay at 64.9. These group means are distinct from task-macro scores. Matching phrases identify a concrete reuse path without assuming stable physical entity IDs across reconstructions.

\subsection{Reuse across task types}
Only one task type triggers revisions in each run; every task is evaluated. All 20 off-diagonal comparisons among absolute distance, size, relative distance, direction, and route planning have positive gains with pointwise intervals above zero. Relative-distance triggers improve direction by 29.7 points [23.8, 35.7]. Counting transfers less: its trigger row has no off-diagonal interval above zero. Counting benefits from relative-distance triggers by 2.3 [1.1, 3.6] and route triggers by 1.2. Room-size questions name no object and do not trigger this experiment. The matrix uses 2,000 paired scene resamples with seed 20261117. Shared objects can explain cross-task reuse; this is not transfer between disjoint referents.

\begin{figure*}[!htbp]
\centering
\includegraphics[width=.88\textwidth]{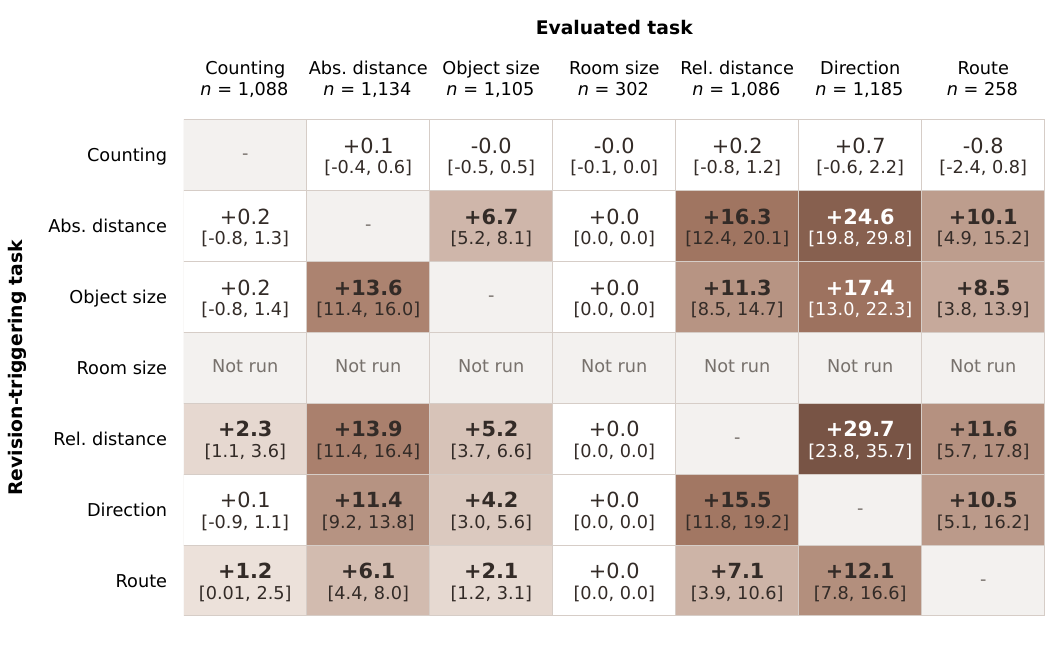}
\caption{\textbf{Reuse across task types.} Rows specify the only task allowed to trigger question-conditioned revision; columns specify the task evaluated from the resulting world. Off-diagonal entries report score gains over the corresponding no-trigger baseline on 380 scenes. Shading marks pointwise 95\% intervals above zero; white cells include zero. Each cell shows its gain and interval from 2,000 paired scene bootstrap resamples. Diagonal cells are omitted because they include the triggering task. The room-size trigger row is not run because those questions name no object.}
\label{fig:supp-task-transfer}
\end{figure*}

\subsection{Continuous updates to one world}
\label{supp:continuous-revision}
We divide questions into four groups and reserve group 0 for evaluation. S1 uses group 1 as triggers; S2 inherits S1 and adds group 2; S3 inherits S2 and adds group 3. Group 0 contains 1,546 questions in 333 scenes; 47 other scenes have none. The reader remains fixed. Controls C2 and C3 instead reconstruct from the saved front end using groups 1--2 and 1--3, respectively.

World records, bindings, and verdict caches persist between rounds; later verdicts replace earlier entries at the same key. Later rounds run phrase verification, recovery, pair checks, size recording, reference resolution, and binding. They skip track-based reconstruction and its inspection sweep. Round-specific new IDs prevent overwriting existing entities. This tests continued world editing under prescribed external triggers, not repetition of the full builder or autonomous target selection.

\begin{table}[!htbp]
\centering
\caption{\textbf{Continuous revision on a fixed evaluation partition.} S2 and S3 inherit the preceding world and use only the next question group as triggers. C2 and C3 rebuild from the saved front end with cumulative triggers. Every condition evaluates the same 1,546 questions. Empty counts unanswered questions. R0 is the saved front end, and S1 equals C1.}
\label{tab:supp-continuous-scores}
\small
\setlength{\tabcolsep}{5pt}
\begin{tabular}{llrr}
\toprule
\rowcolor{suppHeader}
State & Trigger groups & Overall & Empty \\
\midrule
R0 & None & 47.7 & 342 \\
S1 = C1 & 1 & 58.0 & 102 \\
S2 & 1, then 2 & 59.4 & 60 \\
\rowcolor{suppHighlight}
S3 & 1, then 2, then 3 & 61.6 & 38 \\
C2 & 1--2, rebuilt & 59.5 & 62 \\
C3 & 1--3, rebuilt & 61.4 & 42 \\
\bottomrule
\end{tabular}
\end{table}

S2--S1 gains 1.4 points [.5, 2.4]; S3--S2 gains 2.3 [1.1, 3.6]. S3--S1 gains 3.7 [2.2, 5.2], with 120 questions improving and 28 declining. S3--R0 gains 14.0 [11.4, 16.5], with 366 improving and 73 declining. Intervals use 10,000 paired scene bootstrap resamples with seed 20261106. S3 differs from C3 by .2 points [$-.9$, 1.3], and S2 from C2 by $-.1$ [$-1.0$, .7]; these are not equivalence tests.

Gains coexist with regressions. S1--S2 lowers scores on 26 of 1,125 previously positive questions (2.3\%); S2--S3 lowers 13 of 1,163 (1.1\%). S1--S3 lowers 28/1,125 (2.5\%). Positive includes partial numerical credit. \Cref{supp:continuous-audit} separately assesses evidence supporting later-round edits.

\section{Execution Cost}
\label{supp:cost}

Build and revision costs are shared by questions about one scene. These are tool-intensive complete-system comparisons, not equal-compute comparisons with a single direct-video request. The saved main-backbone run records the following stages for 99 scenes on eight RTX~5880 Ada GPUs. Parallel services, batching, cache state, orchestration, and intermediate evaluation reads affect wall-clock time. The table reports collection-level intervals, not per-query latency or a minimal inference budget.

\begin{table}[!htbp]
\centering
\caption{\textbf{Elapsed stages in the recorded main-backbone run.} Times are derived from adjacent stage timestamps for 99 scenes. The tracking-extension interval also includes the subsequent model-service startup. Intermediate evaluation reads and orchestration remain part of the complete execution trace. These collection-level times are not per-question latencies.}
\label{tab:supp-main-runtime}
\small
\setlength{\tabcolsep}{4pt}
\begin{tabularx}{\columnwidth}{Xr}
\toprule
\rowcolor{suppHeader}
Stage & Approximate elapsed time \\
\midrule
Depth and metric scale & 4\,min \\
Initial SAM~3 tracking & 5\,h 47\,min \\
Presence/tracking extension and service startup & 53\,min \\
First inspection sweep & 8\,min \\
Question-driven tracking extension & 4\,min \\
Second inspection sweep & 8\,min \\
Recovery detection & 13\,min \\
\bottomrule
\end{tabularx}
\end{table}

The main run lacks a complete request-level token log, preventing a measured token budget or compute-normalized comparison. An instrumented run on another backbone includes intermediate evaluations and cached perception, so it cannot substitute for this measurement. \Cref{supp:frame-controls} controls visual access only.

\section{Traced Examples}
\label{supp:examples}

\subsection{Binding an existing entity}
In scene \texttt{09c1414f1b}, committed entity \texttt{t11} has label \emph{coffee maker} and stored words \emph{appliance}, \emph{container}, and \emph{coffee maker}. The final binding stage selects it from eight candidates and adds \emph{multi-cooker}. Saved snapshots confirm identical center and extent before and after. The nearest-microwave question changes from C, cutting board, to the annotated D, multi-cooker, because the reader can now retrieve the existing geometry. This verifies one answer transition, not geometric improvement or additional cross-question gains.

\begin{figure}[!htbp]
\centering
\includegraphics[width=\columnwidth]{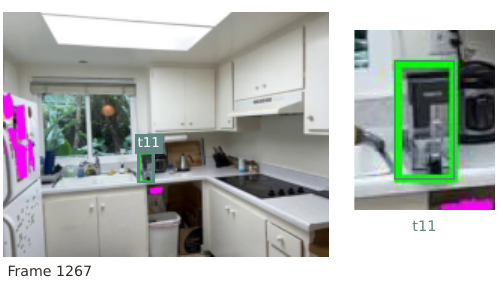}
\caption{\textbf{Adding a binding to an existing entity.} Saved frame 1267 and a detail of \texttt{t11}. The phrase \emph{multi-cooker} is bound to this existing entity. Its geometry is unchanged; the nearest-object answer changes from C to the annotated option D.}
\label{fig:supp-binding-case}
\end{figure}

\subsection{Tracing sketch and video records}
At 50\% sketch removal in scene \texttt{00d42bed}, seven of 13 annotated records remain. Alignment retains these records and admits 63 video entities, which need not be distinct or correctly recovered objects. Replaying the traced readouts reproduces all 11 saved answers.

The floor-lamp size uses sketch record \texttt{g22}: 176.3\,cm, compared with annotated 176\,cm and video-only 120\,cm. The steam-iron--lamp distance combines video record \texttt{p0} with sketch \texttt{g22}: 2.16\,m, compared with annotated 2.6\,m and video-only 2.04\,m. Direction uses video television \texttt{p1}, steam iron \texttt{p0}, and sketch floor lamp \texttt{g22}; both fused and video-only outputs are the annotated option C. This traces the sources of each answer without treating every fused answer as an improvement.

\section{Scope}
\label{supp:scope}

Self-evolution updates persistent world records at inference time with fixed model weights. Our experiments support repeated externally triggered revision and reuse, without establishing autonomous target selection or convergence. Two exploratory self-generated trigger policies add .5 points [.1, .9] and .6 [$-.1$, 1.4] over no question-triggered revision on 380 scenes, with no detected benefit when added to benchmark triggers. Recovery and final-binding operators differ by trigger source in those tests, so the source effect is not isolated.

Better access to world records does not ensure cleaner instance reconstruction. Fragmentation and duplicate entities persist, and incomplete depth or scale cues limit metric geometry. Model-based size estimates can rely on common-sense priors. The original ledger audit is model-assisted; one human reviewer verifies the later-round sample. Edit reliability, QA utility, and physical world accuracy remain distinct evaluation targets.

\end{document}